\documentclass[letterpaper]{article} 
\usepackage{aaai25}  
\usepackage{times}  
\usepackage{helvet}  
\usepackage{courier}  
\usepackage[hyphens]{url}  
\usepackage{graphicx} 
\usepackage{natbib}  
\usepackage{caption} 
\usepackage{algorithm}
\usepackage{algorithmic}

\usepackage{comment}
\usepackage{subcaption}
\usepackage[table,xcdraw]{xcolor}
\usepackage{booktabs}

\usepackage{newfloat}
\usepackage{listings}
\DeclareCaptionStyle{ruled}{labelfont=normalfont,labelsep=colon,strut=off} 
\floatstyle{ruled}
\newfloat{listing}{tb}{lst}{}
\floatname{listing}{Listing}
\title{Accountability Framework for
Healthcare AI Systems: \\Towards Joint Accountability in Decision Making}
\author{
      Prachi Bagave,
    Marcus Westberg,
    Marijn Janssen,
    Aaron Yi Ding
}
\affiliations{
   Delft University of Technology,
The Netherlands \\

p.bagave@tudelft.nl,
m.westberg@tudelft.nl,
M.F.W.H.A.Janssen@tudelft.nl,
Aaron.Ding@tudelft.nl

}

\usepackage{bibentry}

\begin{document}
\nocopyright
\maketitle

\begin{abstract}
AI is transforming the healthcare domain and is increasingly helping practitioners to make health-related decisions. Therefore, accountability becomes a crucial concern for critical AI-driven decisions. Although regulatory bodies, such as the EU Commission, provide guidelines, they are high-level and focus on the “what” that should be done and less on the “how”, creating a knowledge gap for actors. Through an extensive analysis, we found that the term accountability is perceived and dealt with in many different ways, depending on the actor's expertise and domain of work. With increasing concerns about AI accountability issues and the ambiguity around this term, this paper bridges the gap between the “what” and “how” of AI accountability, specifically for AI systems in healthcare. We do this by analysing the concept of accountability, formulating an accountability framework, and providing a three-tier structure for handling various accountability mechanisms. Our accountability framework positions the regulations of healthcare AI systems and the mechanisms adopted by the actors under a consistent accountability regime. Moreover, the three-tier structure guides the actors of the healthcare AI system to categorise the mechanisms based on their conduct.
Through our framework, we advocate that decision-making in healthcare AI holds shared dependencies, where accountability should be dealt with jointly and should foster collaborations. We highlight the role of explainability in instigating communication and information sharing between the actors to further facilitate the collaborative process.

\end{abstract}

%

\section{Introduction}

In critical AI systems, the accountability of AI-assisted decisions is a big concern. The contribution of AI is growing in many fields, given its ability to find correlations from data and make accurate predictions. In the healthcare industry, a critical application area, AI can assist in early disease prediction and aid diagnosis and treatment plans \cite{Jiang2017surveypaperhealthcareAI}. However, since healthcare decisions correspond to an individual’s health, the accountability of such decisions is of the utmost importance. Moreover, when these decisions are made with AI agents, the decision-making process is influenced, and the concerns of accountability become even more complex and ambiguous.

The concerns of AI involved system accountability are even more evident when compared with the conventional healthcare system.
In a traditional healthcare system, the decision-making power is held by the healthcare providers, where patients depend on them to receive the needed treatment and care. For such a system, the accountability mechanisms are well-established and mature \cite{SHORTT2002canadianaccountability,WHOaccountabilityhealthequity}. In such a setting, there are multiple levels of authority-holding organisations \cite{WHOaccountabilityhealthequity}, and public health ministries \cite{NLministryofhealth} to ensure the rights of the patients. However, as AI agents are involved in this decision-making process, the nature of accountability becomes more complicated \cite{Habli2020accountabilityphylosophical} due to the limited transparency of complex models and the uncertainties involved around it. This evolving nature of AI systems in healthcare can alter the benefits and risks of the decision-making process. Because of this, there is a need to understand how this system is evolving, what accountability in such a system would look like, the possible challenges involved, and approaches to solve them.

Many regulatory bodies have provided their recommendations and guidelines to address this accountability aspect.  Regulations, such as the HIPAA and EU GDPR \cite{USHIPAA,EUGDPR}, provide standards for data protection and privacy. The Ethics guidelines for Trustworthy AI and the AI Act \cite{EUethicsguidelinesfortrustworthyAI,EUAIAct2024draftfull} provide guidelines and directives for making the AI trustworthy and accountable against the AI decisions. These recommendations have also been supported by various AI and global health communities \cite{ACMalgotransandacct,whoethicsguidelines}.

Often, such regulatory perspectives and guidelines offer a top-down approach, providing a high-level outlook \cite{EUAIAct2024draftfull,WHOaccountabilityhealthequity,whoethicsguidelines}. However, they are too far from the implementation realities, and there is a clear need for closing this accountability gap by incorporating them in the AI development cycle \cite{2025EthicalandRegulatoryPerspectivesonGenerativeArtificialIntelligenceinPathology,2021AIandDataRightsConsiderationsforUSPolicy}. 
Thus, in this paper, we argue for a bottom-up approach from an actor's perspective, where we highlight mechanisms and tools that the respective actors can utilise for accountability by providing justifications for their actions. Often, regulations are provided from an authoritative perspective and provide mechanisms on "what" should be done. However, the "how" part is left up to the actors to generate actionable items. This gap generates ambiguity in the way accountability is interpreted by the actors, and gives rise to many different interpretations. This paper highlights these interpretations and attempts to provide a bridge between the "what" and the "how", particularly for the Healthcare AI domain.

Moreover, as the black-box nature of machine learning models is considered one of the major bottlenecks for accountability in AI-based systems,  explainability methods have gained a lot of traction. Explainable AI (XAI) is increasingly getting used as one of the popular accountability mechanism, as it can provide insights into the AI decision-making processes \cite{2024PrototypeBasedExplanationstoImproveUnderstandingofUnsupervisedDatasets,2023ExploringCNNandXAIbasedApproachesforAccountableMIDetectionintheContextofIoTenabledEmergencyCommunicationSystems,2022ExplainableDeepLearningBasedEpiretinalMembraneClassificationAnEmpiricalComparisonofSevenInterpretationMethods,2022ExplainableAIforBreastCancerDiagnosisApplicationandUser'sUnderstandabilityPerception,2020ExplainableAIinHealthcare,Mohseni2021AMultidisciplinarySurveyandFramework,Alejandro2020ExplainableArtificialIntelligenceXAIConceptsTaxonomiesOpportunities,Alejandro2020ExplainableArtificialIntelligenceXAIConceptsTaxonomiesOpportunities,TransparencyandaccountabilityinAIdecisionsupport}.This increased transparency in the AI decisions is considered to help the decision-makers take more responsibility for decisions made with the assistance of AI recommendations. However, in this paper, we argue that explainability is only a tool to 
provide accountability for decisions between the agents of a system, and is not a standalone solution for all the accountability issues in AI-based systems.

The critical challenge we aim to tackle in this paper is: 
\textbf{\textit{How to facilitate accountability for decision making in healthcare AI systems?}} 

For this challenge, we explore the following questions:
\begin{enumerate}
    \item In what form does accountability exist in the healthcare AI context, and what are the mechanisms in practice to ensure this accountability?
    \item What are the practical challenges for accountability in healthcare AI systems?
    \item How can explainability address these challenges and contribute towards accountability in decision making?
\end{enumerate}

While finding answers to these questions, we came across accountability from different domains, what they mean in different fields, and this paper provides a structured way to present this interdisciplinary study. We start by giving an overview of the different interpretations of accountability in the Section \ref{section: background}. We observed that there is a lot of ambiguity in the literature around this term, and hence, as a next step towards our framework, we first define what accountability means from an administrative viewpoint. This is executed with the help of some foundational work by Boven \cite{markbovenaccountability}, Roberts \cite{roberts}, Lindberg \cite{Lindbergaccountability2013}, and Novelli et.al \cite{novelli2023accountability} in Section \ref{section:What is accountability}. Further, we build upon the fundamental understanding of accountability to derive a framework for accountability in the healthcare AI system in Section \ref{section:accountability framework}. In this section, we describe the actors of the system, the role of the patients as the principal, and the different authorities for providing compliance mechanisms to ensure accountability. Further, in Section \ref{section: challenges}, we discuss the challenges that this system faces to ensure accountability. Followed by this, we provide some suggestions on how explainability can address these challenges by introducing the three-tier accountability model in Section \ref{sec:three_tier_model}. We also advocate for joint accountability for the decisions taken under the influence of AI assistance as a measure to promote healthy use of AI, and to avoid blaming and scapegoating. Finally, we provide our concluding remarks.

\section{Background} \label{section: background}
The concerns around algorithmic accountability and the need for developing an accountable AI system are at the forefront, and confirmed as a critical topic by various user studies about risks of AI in the medical domain \cite{2023Theapplicationofartificialintelligenceperceptionsfromhealthcareprofessionals,2022SurveyontheperceptionsofUKgastroenterologistsandendoscopiststoartificialintelligence,2020PerceptionsofvirtualprimarycarephysiciansAfocusgroupstudyofmedicalanddatasciencegraduatestudents,2024ImplementationchallengesofartificialintelligenceAIinprimarycarePerspectivesofgeneralpractitionersinLondonUK}.
In the literature, one can find various forms of accountability discussed.
The term "accountability" is often used to convey underlying concepts such as liability, responsibility, transparency, or answerability \cite{JaveroperationalizingAIethics,markbovenaccountability}. As there has been a push to make AI systems more accountable, attempts have been made to quantify this term in technical implementations by introducing reliability, stability, or replicability as the measurable entities \cite{IncorporatingFATandprivacyawareAImodelingapproaches}. Below, we discuss the various forms of accountability found in our literature study, focusing on three key themes: "accountability", "healthcare", and "artificial intelligence".

\textbf{FAT(E) concept}:
AI regulatory agencies provide guidelines for accountability, and group it in acronyms like \textit{FAT or FATE (Fairness, Accountability, Transparency, and Explainability / Ethics)}, highlighting the socio-technical challenges of AI \cite{EUethicsguidelinesfortrustworthyAI,ACMalgotransandacct,DtaprovenanceforFATE,2024BridgingHealthDisparitiesintheDataDrivenWorldofArtificialIntelligenceANarrativeReview,2023EthicalConsiderationsandFairnessintheUseofArtificialIntelligenceforNeuroradiology,2023HarmsfromIncreasinglyAgenticAlgorithmicSystems}. This leads to accountability being discussed in the context of fairness, where a core topic is how AI should ensure accountability measures for upholding individual rights in the data collection process (consent), data security, and ensuring that usage of AI does not impact marginalised groups \cite{2025Harnessingthepowerofartificialintelligencefordiseasesurveillancepurposes}. 

\textbf{Transparency}:
Several other studies, including from the global health organisation, have advocated for \textit{transparency} in the AI development process to address the accountability issue raised by multiple actors working together under different jurisdictions \cite{2024Researchethicsandartificialintelligenceforglobalhealthperspectivesfromtheglobalforumonbioethicsinresearch,2024UNVEILINGTHEBLACKBOXBRINGINGALGORITHMICTRANSPARENCYTOAI,2024BridgingHealthDisparitiesintheDataDrivenWorldofArtificialIntelligenceANarrativeReview,2024AStudyonEthicalConsiderationsinAutomatedLungUltrasoundAnalysis}. 
S. Zyryanov (\citeyear{2023ArtificialIntelligenceasaMeansofRosselhoznadzorProblemsandProspects}), however, argues that in some cases, this might not be possible due to the presence of non-disclosure agreements, trade secrets, or technical complexity. This also opens up the discussion for the trade-off between having transparency for accountability and adhering to data privacy concerns. On the other hand, Bickley and Torgler (\cite{2023Cognitivearchitecturesforartificialintelligenceethics}) argue that transparency is a necessary condition for accountability, but not sufficient. For confidential or data with intellectual properties, they coin the term opaque transparency, where insights are presented without full disclosure.

\textbf{Responsibility}:
Accountability is also closely related to \textit{responsibility}, where it is needed to understand the responsible party for the algorithmic judgements \cite{2023ArtificialintelligenceAIoraugmentedintelligenceHowbigdataandAIaretransforminghealthcareChallengesandopportunities,2024ProposinganAIPassportasaMitigatingActionofRiskAssociatedtoArtificialIntelligenceinHealthcare,2020PerceptionsofvirtualprimarycarephysiciansAfocusgroupstudyofmedicalanddatasciencegraduatestudents,2024AnalyzingEthicalDilemmasinAIAssistedDiagnosticsandTreatmentDecisionsforPatientSafety}.

There is an ongoing discussion about whether to address the accountability gap by clearly \textit{distributing} responsibilities among the actors \cite{2024ProposinganAIPassportasaMitigatingActionofRiskAssociatedtoArtificialIntelligenceinHealthcare, 2024ArtificialIntelligenceAlgorithmsinCardiovascularMedicineAnAttainablePromisetoImprovePatientOutcomesoranInaccessibleInvestment}, or to handle it by \textit{sharing} them \cite{BELL2011519,2025EthicalandRegulatoryPerspectivesonGenerativeArtificialIntelligenceinPathology}. 
Moreover, C. Ewuoso (\citeyear{2023BlackboxproblemandAfricanviewsoftrust}) highlights the dilemmas of accountability in the healthcare AI scenario. They argue that healthcare professionals cannot be completely accountable for decisions taken with AI assistance, as the black-box AI poses problems. Under such circumstances, it is difficult to achieve the two critical requirements for accountability, i.e. the decision-maker having control over the decisions and being aware of the consequences thereof.
In contrast to the accountability of human experts, F. Karimov (\citeyear{2024TheVariousTechnicalSecuritytobeTakenintoConsiderationsforImplementationofArtificialIntelligenceinVariousApplications}) discusses algorithmic accountability, described as the difficulty in assigning responsibility for AI decisions. As an algorithm cannot be held accountable for its decisions, clear human oversight is necessary for such scenarios. Along similar lines, A. Chan et al. (\citeyear{2023HarmsfromIncreasinglyAgenticAlgorithmicSystems}) argue about the increasing agency of AI systems, and yet assert to pass the moral responsibility (and thus accountability) to human developers.



\subsection{Addressing Accountability through Technical Implementations}

The need for accountable systems has led to a number of technical solutions that reflect this need. From an organisational point of view in the AI industry, accountability may commonly be understood as the ability to ensure quality, responsibility, and protection over the data, algorithms and results of the AI in question \cite{JaveroperationalizingAIethics}. Throughout the literature, a number of solutions claim to contribute to accountability and propose solutions as a step closer to addressing this issue. Table \ref{tab:overview of tech implementation} provides an overview of the technologies that claim to address this issue with their technical implementations and their implicit arguments to contribute to accountability.

From the Table \ref{tab:overview of tech implementation}, five main technologies are observed to contribute to accountable systems: data quality mechanisms, federated learning, blockchain, XAI, and humans in the loop. The table also provides underlying concepts beneath their arguments, which relate to accountability. Various mechanisms for ensuring data quality have been discussed, such as data trusts \cite{data_trust_canada_2020}, data provenance \cite{DtaprovenanceforFATE} and data logger\cite{2023DataRecordingforResponsibleRobotics}. Ensuring the data quality in terms of biases, trustworthy sources and verified labelling \cite{2022TowardsTransparencyinDermatologyImageDatasetswithSkinToneAnnotationsbyExpertsCrowdsandanAlgorithm} processes ensures that the data models are built on fair, trustworthy data. Using data logger and data provenance methods promotes data traceability to ensure data quality \cite{DtaprovenanceforFATE,2023DataRecordingforResponsibleRobotics}. Additionally, data trusts provide safe and secure data sharing arrangements, necessary specifically for health data \cite{data_trust_canada_2020}. 

It has been argued that XAI has the potential to provide insights into the working of AI algorithms and thus can help in accountability justifications \cite{2024PrototypeBasedExplanationstoImproveUnderstandingofUnsupervisedDatasets,2023ExploringCNNandXAIbasedApproachesforAccountableMIDetectionintheContextofIoTenabledEmergencyCommunicationSystems,2022ExplainableDeepLearningBasedEpiretinalMembraneClassificationAnEmpiricalComparisonofSevenInterpretationMethods,2022ExplainableAIforBreastCancerDiagnosisApplicationandUser'sUnderstandabilityPerception,2020ExplainableAIinHealthcare}. However, truthfulness, human interpretability, and causality are some properties that help make them suitable for this \cite{2024Analysisandinterpretabilityofmachinelearningmodelstoclassifythyroiddisease,2024CausalInferenceinAIBasedDecisionSupportBeyondCorrelationtoCausation}. Federated Learning, another widely used technique in distributed environments, supports privacy protection mechanisms and provides resilience against data attacks. This contributes to the accountability issues related to data protection and privacy preservation as mentioned in the HIPAA regulations (\citeyear{USHIPAA}) or the EU GDPR (\citeyear{EUGDPR} \cite{2025InformationModelingTechniquetoDecipherResearchTrendsofFederatedLearninginHealthcare}). On similar lines, Blockchain and advanced blockchain also provide decentralised data storage, and additionally, ensure data authenticity and accountability in the hands of the people involved. Thus, it also contributes to the accountability of data, ensuring its security, accuracy, traceability and transparency in data sharing \cite{2025BlockchainDrivenSupplyChainFinanceforPublicHealthcareinIndiaEnhancingFinancialResilienceinPublicHealthSystems,2023ABlockchainbasedModelforMaternalHealthInformationExchangeandPredictionofHealthRisksusingMachineLearning,2020SmarteHealthSecurityandSafetyMonitoringwithMachineLearningServices}.

\begin{table*}
\centering

\begin{tabular}{@{}|p{0.15 \textwidth} p{0.4 \textwidth} p{0.15 \textwidth} p{0.2 \textwidth}|@{}}
\toprule

\textbf{Implemented   Technology}     
& \textbf{Arguments contributing to accountability}                                           & \textbf{Underlying concept relating to accountability}                                     & \textbf{References} \\

\midrule
Data Trust
& Data trusts are mechanisms that enable fair, safe and equitable sharing of data
&  data safety \& security
&  \cite{data_trust_canada_2020}
\\
Data Provenance 
& Data provenance methods facilitate record keeping about the origin and processing of data. This ensures data traceability, essential for maintaining data quality, and in turn ensures the fair and transparent use of data. 
&Data Traceability
& \cite{DtaprovenanceforFATE}\\

Data Logger 
& Data recorder is a data logging technique used in robotics for ex-post investigation (similar to flight data recorders) 
&Data Traceability
& \cite{2023DataRecordingforResponsibleRobotics}\\
\\
\midrule
Federated   learning                  
& Federated learning offers a privacy-preserving, distributed approach to learning, where multiple connected devices collaboratively train a model, without sharing their data. This is in accordance with the HIPAA regulations imposing resistance to attacks, algorithmic accountability, and data security. 
& Data privacy and resilience to attacks.
& \cite{2025InformationModelingTechniquetoDecipherResearchTrendsofFederatedLearninginHealthcare}
\\ \midrule

Blockchain 
& Blockchain facilitates decentralised data storage and secures it against data attacks. The computations are transparent, and all involved parties are accountable for their actions, fostering traceability, authenticity, and data accuracy.                       
& Data security, transparency, responsibility, authenticity, traceability, and data accuracy 
& \cite{2025BlockchainDrivenSupplyChainFinanceforPublicHealthcareinIndiaEnhancingFinancialResilienceinPublicHealthSystems,2023ABlockchainbasedModelforMaternalHealthInformationExchangeandPredictionofHealthRisksusingMachineLearning}                    
\\ 
Advanced   blockchain                 
& Advanced blockchain enables virtualised logging service, automatic scaling and resiliency for e-health monitoring solutions. This also helps auditing processes                               & Traceability and resilience                &  \cite{2020SmarteHealthSecurityandSafetyMonitoringwithMachineLearningServices}
\\
\midrule

XAI
& XAI enables understanding AI decision-making processes by providing insights into how and why specific outcomes are reached. This transparency offers a solution to the accountability issue.   
& Transparency                                                                              
& \cite{2024PrototypeBasedExplanationstoImproveUnderstandingofUnsupervisedDatasets,2023ExploringCNNandXAIbasedApproachesforAccountableMIDetectionintheContextofIoTenabledEmergencyCommunicationSystems,2022ExplainableDeepLearningBasedEpiretinalMembraneClassificationAnEmpiricalComparisonofSevenInterpretationMethods,2022ExplainableAIforBreastCancerDiagnosisApplicationandUser'sUnderstandabilityPerception,2020ExplainableAIinHealthcare,Mohseni2021AMultidisciplinarySurveyandFramework,Alejandro2020ExplainableArtificialIntelligenceXAIConceptsTaxonomiesOpportunities}
\\ 
XAI and   user studies
& Verify the correctness of explanations by mapping to the expert opinions or the users. If explanations align well with the expert opinion, they are more likely to use them to justify their accountability.
& Expert-aligned explanations / truthful explanation   
& \cite{2024Analysisandinterpretabilityofmachinelearningmodelstoclassifythyroiddisease}                   
\\ 
Causal inference
& Causal explanations are more likely to be truthful and are suitable for providing justifications for AI-assisted decisions.                                                     

& Truthful explanations                                      & \cite{2024CausalInferenceinAIBasedDecisionSupportBeyondCorrelationtoCausation}
\\
Counterfactual adversarial examples 
& Training a model to account for adversarial examples can ensure risk management for critical scenarios in the medical domain.                                                       
& Resiliency                                                 & \cite{2019Towardanunderstandingofadversarialexamplesinclinicaltrials}                     
\\ 
\midrule
Human in the loop                   
& Human in the loop ensures safety and alleviates the ambiguity around accountability.  Compliant with USAID’s guide to machine learning.
& Human oversight                                                                              &   \cite{2023TransformingRapidDiagnosticTestsintoTrustedDiagnosticToolsinLMICusingAI}                  

\\ \bottomrule
\end{tabular}%
\caption{Technical implementations addressing accountability, their arguments for the contributions, and the underlying concept relating to accountability
}
\label{tab:overview of tech implementation}
\end{table*}
\subsection{AI Regulations for Accountability in Healthcare AI Systems}
\label{section:AI regulations}
In the EU, the Ethics guidelines for Trustworthy AI (\citeyear{EUethicsguidelinesfortrustworthyAI}), and the AI act (\citeyear{EUAIAct2024draftfull}) provide high-level guidelines and compliance mechanisms for accountability measures in AI systems. According to the AI Act (\citeyear{EUAIAct2024draftfull}), healthcare systems are classified as critical systems and impose strict compliance with them. Additionally, for regulating the data, the Data Act (\citeyear{Data_act}), the European Data Governance Act (\citeyear{Data_governance_act}), and the EU GDPR (\citeyear{EUGDPR}) are also relevant. 

\subsubsection{Ethics Guidelines for Trustworthy AI}

In 2019, the Ethics Guidelines for Trustworthy AI (\citeyear{EUethicsguidelinesfortrustworthyAI}) provided seven non-binding principles for trustworthy AI: human agency and oversight; technical robustness and safety; privacy and data
governance; transparency; diversity, non-discrimination and fairness; societal and
environmental well-being and accountability. Further, it also provides the following requirements for AI accountability:
\begin{itemize}
    \item   \textit{Auditability}: Allowing audits by internal or external auditors to assess data, algorithms, and design processes.
    \item  \textit{Minimisation and reporting of negative impacts}: Identifying, assessing, documenting and minimising potential risks, and allowing external parties to report harmful impacts.
    \item   \textit{Trade-offs}: The trade-offs faced during the implementation of these requirements must be acknowledged and evaluated against ethical principles.
    \item   \textit{Redress}: In case of adverse impacts, mechanisms for redress should be ensured.
\end{itemize}

These guidelines serve as the basis for providing many regulations, as also observed in the AI ACT (\citeyear{EUAIAct2024draftfull}). However, the enforcement mechanisms became concrete, as the principles provide non-binding guidelines compared to the AI ACT, which contains compliance mechanisms for high-risk systems.

\subsubsection{AI Act}
The AI Act (\citeyear{EUAIAct2024draftfull}) classifies AI systems according to their risks, and provides regulations for AI systems in three levels of risk classification: general purpose, high risk, and prohibited AI systems. According to this classification, the healthcare AI systems are considered as high risk systems.
This obligates the developers of AI systems to ensure the following:
\begin{itemize}
    \item   Establish a risk management system that is continuously maintained, documented, and implemented, where the reasonably foreseeable risks are identified, estimated, and mitigated in the system.
    \item   Conduct data governance, ensuring that datasets are relevant, sufficiently representative, error-free and complete according to the intended purpose.
    \item   Draw up technical documentation to demonstrate compliance, and provide authorities with the information to assess that compliance.
    \item   Maintain records of events relevant for identifying risks and substantial modifications throughout the system’s lifecycle.
    \item   Provide sufficient transparency for deployers and provide clear, complete and concise instructions for the users of the system.
    \item   Enable ways for human oversight, so that natural persons can intervene to prevent and minimise risks to health, safety, or fundamental rights.
    \item   Maintain appropriate levels of accuracy, robustness, and cybersecurity, in cooperation with relevant stakeholders and benchmarking authorities.
    \item   Establish a quality management system to ensure compliance with these regulations.
\end{itemize}


\section{What Is Accountability?}
\label{section:What is accountability}

Accountability is a broad term, connected to many underlying concepts as "answerability", "responsibility", "transparency", "responsiveness" and "integrity" \cite{JaveroperationalizingAIethics,markbovenaccountability}. This creates ambiguity, as it occurs in various forms.
For example, the formal mechanisms such as the judicial or legislative processes have high-order compliance mechanisms in place, and the informal mechanisms such as in the executive, or societal procedures, are derived from the broad-level professional work culture, and societal ethical concerns \cite{markbovenaccountability}. The nature of accountability also depends on various other factors like the presence of an authoritative control, the the degree of compliance, the nature of actors and conduct \cite{roberts,markbovenaccountability}, and whether it is forward or backward looking \cite{novelli2023accountability}. 
Despite such a wide variation and complex nature of existence, the following definitions provide a general understanding of accountability :

\begin{quote}
\textit{Accountability is the obligation to answer for one's actions or inaction and to be responsible for their consequences \cite{roberts}.}  
\end{quote}

\begin{figure}[!b]

\includegraphics[width=\columnwidth]{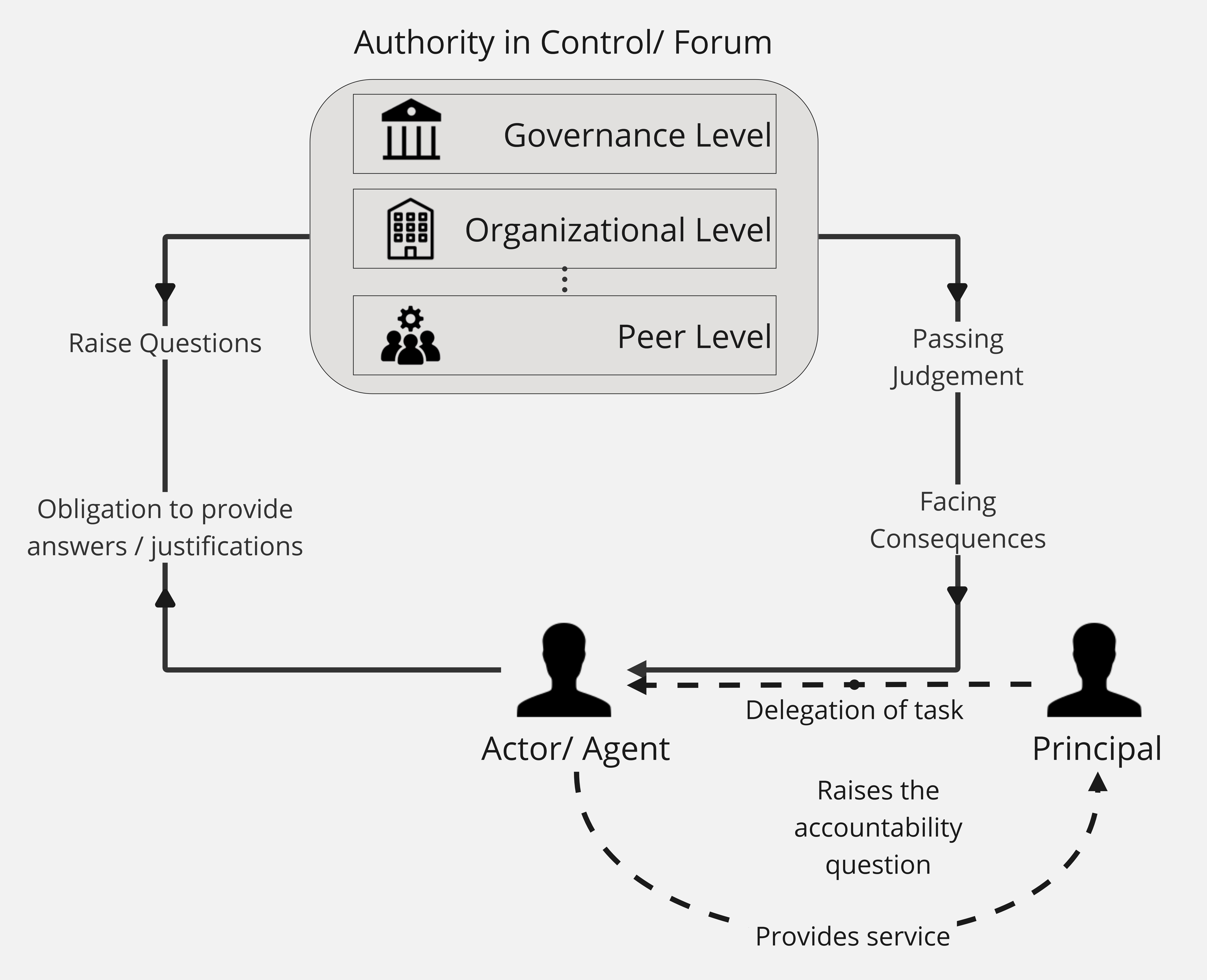}\centering
\caption{A general understanding of accountability in a healthcare AI context, showing the relationship between the actor, authorities in control, and the implicit role of principal. Inspired by the accountability definitions \cite{markbovenaccountability,Lindbergaccountability2013,novelli2023accountability}.}
\label{fig:conceptual_acc_simple}
\end{figure}

Accountability is commonly understood as the obligation of providing justifications for one's actions (or inactions) \cite{roberts}. This obligation can be of a person, a group of people or an organisation, and are referred to as the \textit{actors} in the definitions (refer to Fig. \ref{fig:conceptual_acc_simple}).

\begin{quote}
\textit{Accountability is a relationship between an actor and a forum, in which the actor has an obligation to explain and to justify his or her conduct, the forum can pose questions and pass judgment, and the actor may face consequences }\cite{markbovenaccountability}.
\end{quote}
 Additionally, Boven \cite{markbovenaccountability} provides the three elements of accountability as the actor, the forum (also referred to as the authority in control), and the relationship between them, where the actor has the obligation to provide answers and has to face consequences posed by the forum based on their judgment. Boven's work highlights an explicit presence of an authority to provide compliance mechanisms and to pass judgments as observed in a formal setting. In contrast, Robert \cite{roberts}, poses it as a virtue of the actor, regardless of the presence of the forum. The differences in definitions arise due to the different forms of accountability discussed in their work.

\begin{quote}
\textit{When decision-making power is transferred from a principal (e.g. the citizens) to an agent (e.g., the government), there must be a mechanism in place for holding the agent accountable for their decisions and tools for sanctioning \cite{Lindbergaccountability2013}.}
 
\end{quote}




Moreover, according to Lindberg \cite{Lindbergaccountability2013}, the question of accountability arises in a principal-agent relationship, where an agent A is delegated with certain powers by the principal P, in order to perform specific tasks and responsibilities, and the mechanism of accountability is needed to keep this power in check. However, their work does not reflect upon the presence of the forum, or any other authority in control. As observed, their discussions are heavily based on political science, where the government serves as agent A, and citizens as principal P, and by electing the government, the citizens delegate the power of governance to the democratic body. According to their work, the principal is the delegator of power as well as the one holding rights to sanction the agent. 
 
Novelli \cite{novelli2023accountability} builds on the work by Bovens \cite{markbovenaccountability}, Lindberg \cite{Lindbergaccountability2013}, and Mulgan \cite{Mulgan2000AccountabilityAE}, and redefines accountability in the context of Artificial Intelligence. Their work distinguishes between the principal, who delegates the responsibilities, and the forum, which is responsible for asking justifications and providing sanctions. We build our work on this definition of accountability, as in a healthcare AI context, these tasks are also handled by different parties. 

We distinguish between the principal, the agent, and the forum, as these roles are performed by different organisations in a healthcare AI system. Figure \ref{fig:conceptual_acc_simple} explains this with a simplified conceptual illustration.
In our given scenario, the patients act as the principal, as they delegate the power of taking care of their health into the hands of the experts (healthcare experts, or more actors, as in the case of healthcare AI systems). These are the people who are affected by the decisions taken by the experts, and can demand justice for the decisions taken. However, there is a different set of institutes (the forum) to ensure the work done by the agents is justified, can ask for clarifications, and pose their judgments. In a healthcare AI system, authority (or the forum) may be at various levels, as there are multiple levels of accountability in the system. We continue on this understanding of accountability, and derive a detailed framework for Healthcare AI in Section \ref{section:accountability framework}.

Accountability can also be of different types, such as legal, political, administrative, bureaucratic, professional, social,  or even moral \cite{markbovenaccountability,roberts}.
It may even be a combination of these types. However, the enforcement methods, the presence of standardised methods, and the degree of implications on the agents all play a crucial role in shaping it. As Roberts \cite{roberts} discusses, being responsible indicates a person's agency and authorship over their actions. At the same time, it makes them answerable to a higher authority. This creates tension, as there exists a paradox between personal authorship and the obligation of accountability. In addition, scarcity of information, unclear responsibilities, and loose enforcement methods are some additional situations that give rise to scapegoating conditions and the deterioration of personal and moral responsibility. The challenge is to find a balance between agency, obligation, and accountability.

\section{A Framework for Accountability in Healthcare AI Systems}
\vspace*{5pt}
\label{section:accountability framework}

\begin{figure*}[h!]
    \centering
    \includegraphics[width=1\textwidth]{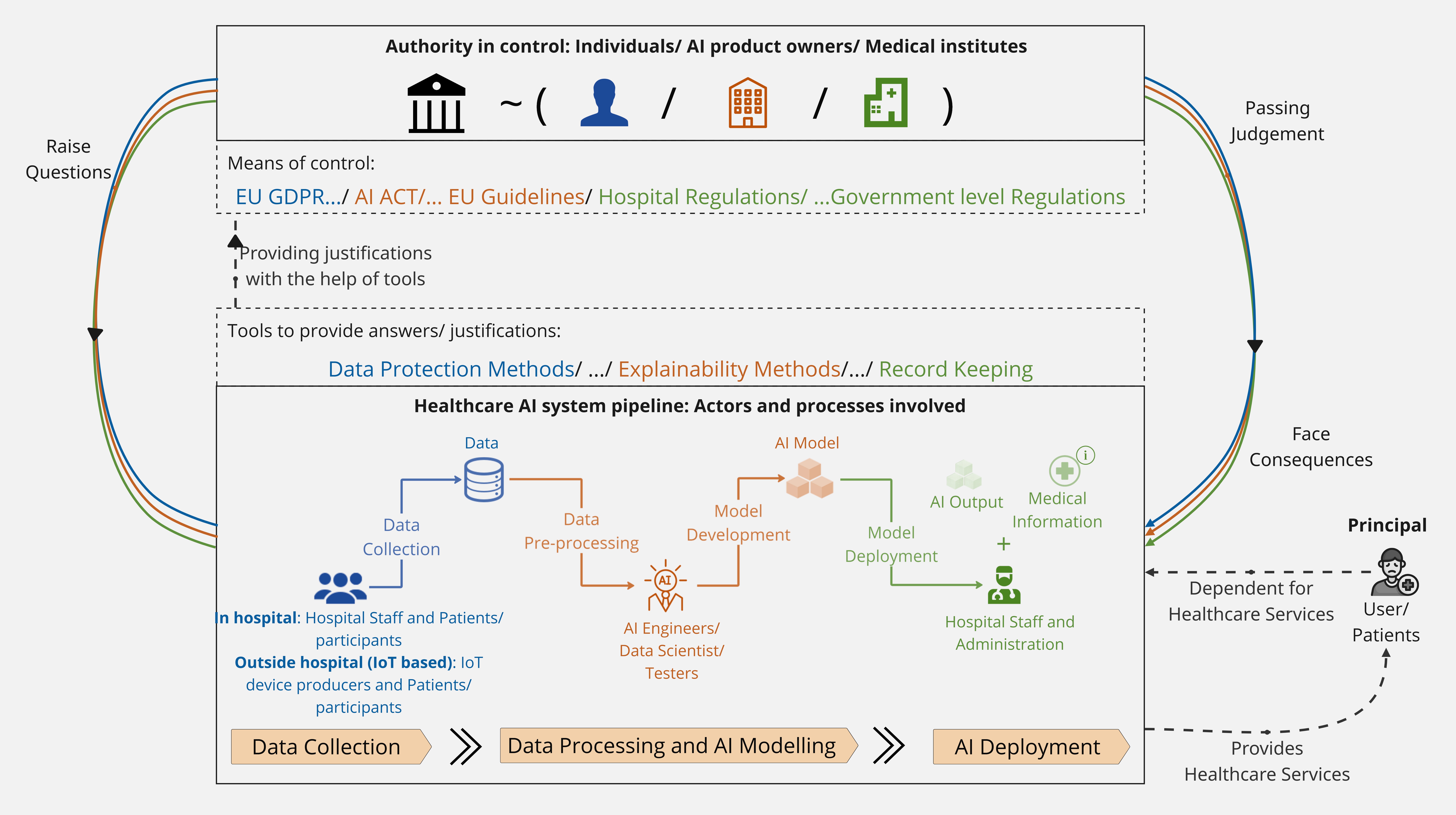}
    \caption{A framework for accountability in Healthcare AI systems, showing the various organisations acting as authority in control and the roles of the actors. The \textit{means of control} highlights the regulatiosn and guidelines imposed by the authority in control to ensure accountability, and the \textit{tools to provide answers and justifications} reflect various mechanisms in practice by the actors. The organisations are colour coded: blue for the individuals and organisations providing data, orange for AI product owners, and green for healthcare institutes.}
    \label{fig:acc_for_HC_AI}
\end{figure*}

We extend our understanding of accountability from the foundational definitions and apply them to a healthcare AI setting. We expand on the three important elements of accountability: the actors, the authority in control, and the principal. As depicted in Figure \ref{fig:acc_for_HC_AI}, the healthcare AI system consists of all the actors present in the pipeline of AI development and deployment. This process starts from the data collection process, further into the development, and finally into the deployment phase. The various institutions and authorities governing these actors form the authority in control of the whole ecosystem. As the patients are the delegators, they act as the principal in the framework. Our framework further highlights "what" the governing bodies impose to enforce the accountability mechanisms. These typically occur as compliance mechanisms and regulations providing the directives and principles for the actors. Further, the measures taken by the actors serve the purpose of justifications of their actions.

In our framework, we consider the complete Healthcare AI system as one complex set of actors, acting as agents, to provide health benefits to the patients, acting as the principal. The rationale behind this is that the patients do not distinguish between the various actors present in this system. In addition, each step in the pipeline serves as a stepping stone for the next one. Thus, even though the actors work independently, the system impacts the patient as a single entity. As a result, the healthcare industry is transforming as a smart industry along with the advancements in AI \cite{2023Callfortheresponsibleartificialintelligenceinthehealthcare}. This emphasises the need to consider the system as an interdependent set of actors working together. 

The healthcare AI system (acting as agent A) consists of a series of interdependent processes, namely, data collection, data processing, AI modelling, and model deployment. These processes may involve one or more organisations. The Figure \ref{fig:acc_for_HC_AI} further highlights the organisations and actors in a colour-coded fashion. For example, the data may be collected in the hospital, as in the case of Electronic Health Data, or outside the hospital setting by IoT devices. The data processing and modelling are typically the responsibility of a team of data scientists and engineers, and finally, the system is deployed in the application scenario to make health-related decisions, where the healthcare experts are involved in the decision-making process. Thus, the whole system contains a range of experts, including lay participants using the IoT devices, technical experts on AI, and healthcare experts. 

Presently, depending on the types of organisations, corresponding governing authorities act as the authority in control. For example, the healthcare experts are governed by the hospital norms, and the AI experts by their corresponding organisations. Although the various stages are interlinked and interdependent on each other, there is no governing body which oversees the whole system. This is one of the major issues faced by the AI industries, as they have little to no control over the application domain, sometimes lack the expertise in that area, and further face model deployment issues due to the dynamic nature of the real-time data.

In Figure \ref{fig:acc_for_HC_AI}, we also highlight some of the regulations, guidelines and recommendations from the regulatory boards and institutes in authority as \textit{the means of control}. These constitute the standards for compliance, one of the important features discussed by Novelli \cite{novelli2023accountability}. For the Healthcare AI industry, these may consist of the EU GDPR \cite{EUGDPR} for regulating the use and protection of data, the AI Act \cite{EUAIAct2024draftfull} and EU Guidelines \cite{EUethicsguidelinesfortrustworthyAI} for enforcing the safe use of AI systems, and the hospital and other government-level regulations for ensuring public health. Although the regulatory bodies may refer to these documents when raising accountability questions, there are different mechanisms adopted by the actors to provide their justifications, referred to as \textit{tools to provide justifications} in the figure. Some examples of such tools as observed from the literature are data provenance methods for providing the quality of data \cite{DtaprovenanceforFATE}, explainability methods for justifying the AI decisions \cite{2024PrototypeBasedExplanationstoImproveUnderstandingofUnsupervisedDatasets,2023ExploringCNNandXAIbasedApproachesforAccountableMIDetectionintheContextofIoTenabledEmergencyCommunicationSystems,2022ExplainableDeepLearningBasedEpiretinalMembraneClassificationAnEmpiricalComparisonofSevenInterpretationMethods,2022ExplainableAIforBreastCancerDiagnosisApplicationandUser'sUnderstandabilityPerception,2020ExplainableAIinHealthcare,Mohseni2021AMultidisciplinarySurveyandFramework,Alejandro2020ExplainableArtificialIntelligenceXAIConceptsTaxonomiesOpportunities}, and record-keeping \cite{cobbereviewableautomateddecisionmaking} methods to provide process-level justifications.

As observed from our framework, there could be multiple levels of authority in control from the government to the organisational level, different levels of enforcement of standards, and various mechanisms to provide justifications for accountability in the healthcare AI industry. As observed in the background section, a number of different techniques are used while referring to accountability methods. However, the contribution of these mechanisms to accountability justifications is not always clear, leaving space for multiple accountability interpretations. Thus, our framework provides a baseline that may be used to understand how and why a particular mechanism may contribute to accountability measures.


\section{Challenges for Ensuring Accountability in Healthcare AI Systems}
\label{section: challenges}
Although we have corresponding governing authorities for different organisations present in the system, we still lack an overarching accountability mechanism, proper inter-organisation communication, and mechanisms for joint accountability. In this section, we discuss the challenges to provide accountability for the Healthcare AI system as observed from our framework. 


\subsection{Multiple Organisations with Independent Authority in Control and Unclear Handover Processes}

From Figure \ref{fig:acc_for_HC_AI}, it can be derived that a Healthcare AI system may have 2-3 institutions involved in the process: the data collection group (or individuals), the AI organisations, and the hospital staff. While the hospital industry is the oldest, with reporting systems and governing regulations in place, the regulations for the AI industry and data usage are still fresh and constantly updated \cite{Data_act,EUAIAct2024draftfull}. This causes the problem of many hands, and the problem of many eyes \cite{markbovenaccountability,wieringa}, as there are many independent organisations present, governed by different regulatory bodies. Additionally. There is a lack of central control, missing information from one process level to another, and a lack of communication standards. As a result, this provides opportunities for scapegoating, and accountability gaps \cite{roberts}.



\subsection{Unclear Accountability Structure for Shared Dependencies}
Although the system is interdependent, the accountability of the final decision is often borne by the healthcare professionals. They always carry the risk, as the responsibility of whether or not to act as per the AI decision also stays with them \cite{2024ImplementationchallengesofartificialintelligenceAIinprimarycarePerspectivesofgeneralpractitionersinLondonUK}. 
In an interdependent system, however, the final decision is influenced by multiple factors. Thus, it can be called into question whether accountability for the final decision can really be attributed to a singular party. Although each individual party may be held accountable for the process they are responsible for, modern healthcare increasingly views accountability as shared between groups of healthcare providers \cite{BELL2011519, 2025EthicalandRegulatoryPerspectivesonGenerativeArtificialIntelligenceinPathology}. Complex healthcare systems cause difficulty in ascertaining the boundaries of individual accountability, if it can be determined at all. As such, improving frameworks for joint accountability is an important challenge to overcome.

\subsection{Interdisciplinary Miscommunication}
Even though the final decision is made by the practitioner, their decision is a byproduct of the AI information handled by many other people in the AI pipeline. As there is a lack of handover processes within the organisations, the presence of multiple disciplines makes the task even difficult. To add to it, the black box nature of the machine learning models makes this even more challenging. This occurs because of the (necessarily) complex nature of the AI algorithms used for decisions, and the inability to fully explain the reasoning behind a decision. Opaqueness contributes to a degree of uncertainty about the correctness of AI outputs. Explainability has been used as a means to address this issue. However, what and how to explain remains a topic of research interest. One reason behind this is that we often lack
feedback processes or a continuous development process to
ensure the information flows within organisations and different disciplines.


\section{Three-Tier Accountability Structure:
A Bottoms-Up Approach}
\label{sec:three_tier_model}

\begin{figure*}[h!]
\centering
\includegraphics[width=\textwidth]{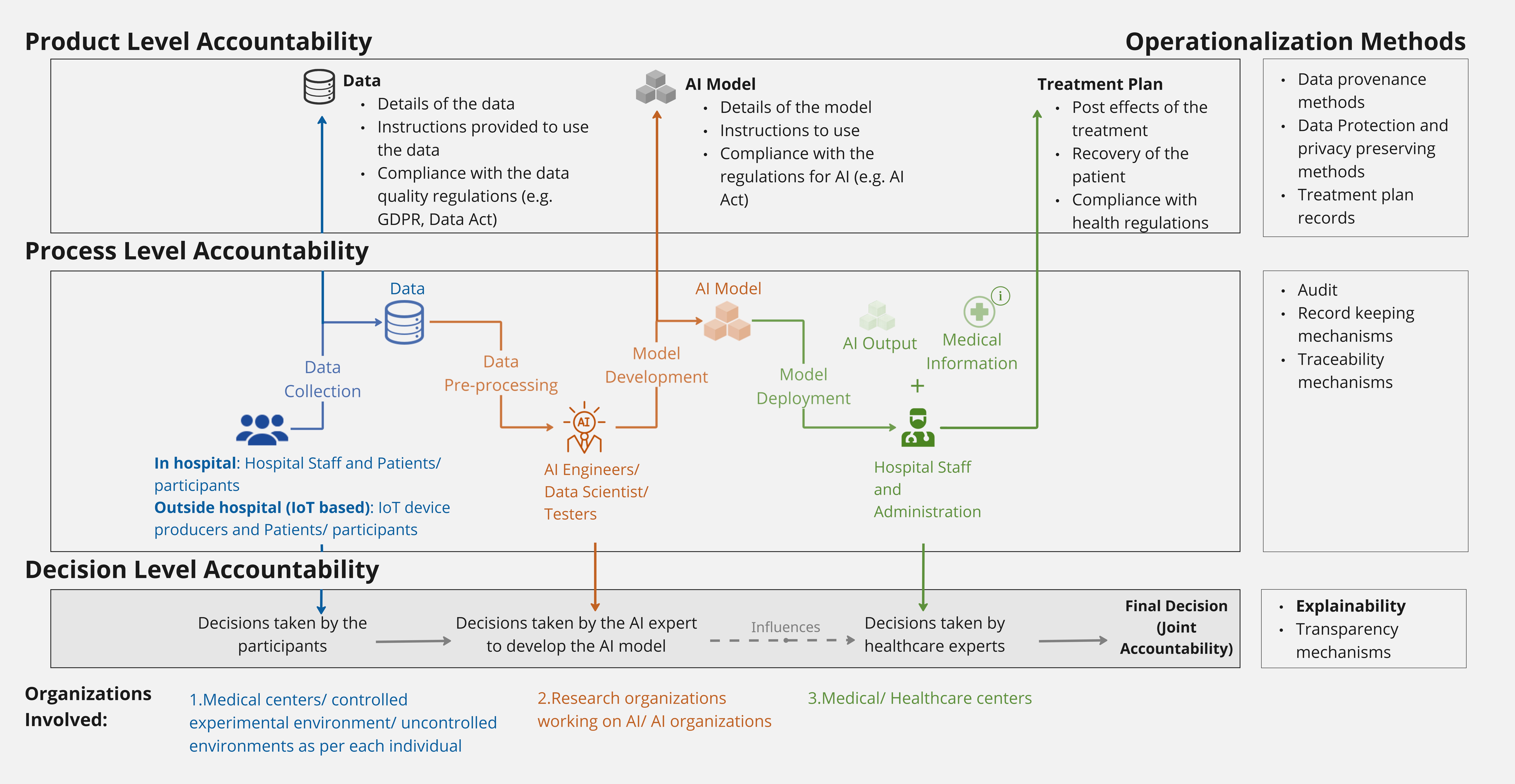}
\caption{A three-tier accountability structure for the actors of the healthcare systems based on their conduct. On the right column, some operationalisation methods are mapped to these tiers, highlighting \textit{explainability} contributing to the decision-level accountability.}
\label{fig:Acc_across_organizations}
\end{figure*}

In this section, we provide a three-tier accountability structure for the actors of healthcare AI systems. We classify this according to Boven's classification of accountability based on the aspect of conduct\cite{markbovenaccountability}. According to their definition of accountability, an actor is obliged to justify their conduct, and this conduct may have various aspects such as financial, procedural, product, and many more. On similar lines, as depicted in Figure \ref{fig:Acc_across_organizations}, we classify the conduct of our actors of the healthcare AI systems into three categories: product, process, and decision level accountability. We further map some mechanisms in practice to these levels of accountability, to position them in our framework.

\subsubsection{Product Level Accountability:} Product level accountability corresponds to the three different products from the actors of our healthcare AI system: the data, the AI model, and the treatment plan from the healthcare practitioner. Each of these is distinctive in its nature and expertise, and an independent governing body may ensure their quality and safety of the product. For e.g. the data regulations such as the EU GDPR \cite{EUGDPR}, the Data Act \cite{Data_act}, and the US HIPAA \cite{USHIPAA} ensure the governance and accountability measures of the data. For the AI models, the emerging AI Act \cite{EUAIAct2024draftfull} ensures performance evaluations, risk assessments and contractual obligations and promotes better-performing models with risk minimisation and redressing mechanisms. Finally, the treatment plans and medical procedures are well governed by the healthcare departments. For commercially viable products, ensuring product quality also affects financial as well as its organisational reputation in the market, leading to corporate accountability where the whole organisation acts as a responsible actor \cite{markbovenaccountability}. Such factors lead to strong motivation for ensuring the product-level accountability, which can also be observed from the various methods for operationalisation in practice.

\subsubsection{Process Level Accountability:} Process level accountability ensures that the processes for product development and deployment ensure safety, risk minimisation, and benevolence of the products. Traces of process-level accountability have been observed in the AI Act \cite{EUAIAct2024draftfull}. We also observed work advocating for record keeping and traceability as accountability measures for AI-assisted systems \cite{Krolljoshuatraceability}. Jennifer Cobbe et. al. \cite{cobbereviewableautomateddecisionmaking}, also provide a framework for accountability in algorithmic systems, where reviewability across the AI life cycle is promoted. These measures promote transparency in the process of AI development rather than just the product. This may include, for e.g. logs of decisions made during the processes, details of the operating procedures, iterations of the development process and other relevant details for bringing meaningful transparency. It also provides details on the steps involved during the process of development and the considerations of trade-offs made during this time \cite{whoethicsguidelines}. The auditing mechanism is one well-known method to operationalise this.

\subsubsection{Decision Level (Joint) Accountability:} Although at the product level and the process level, each organisation has an individual role to play, the same cannot be stated about the accountability for the decisions being made. Especially, for the decisions concerning the treatment plans of patients, even though the final decision is being made by the hospital staff, the accountability does not necessarily rely solely on them. From the literature, we understand that decision-making may be affected by two main factors: the control condition and the epistemic condition \cite{Habli2020accountabilityphylosophical}. The control condition defines how much control the decision maker has over the decisions, and this can vary depending on whether the AI system is fully automated or only used for assistance. Even when the AI is only used for assistance, we argue that the involvement requires some level of training, cognitive thinking, and time allocation, and such conditions have their indirect effects on the decisions. For epistemic conditions, which require sufficient understanding of the decision and its consequences, AI manifests a clear bottleneck due to its black box nature. Due to such influences on the final decisions, we argue that the accountability of the final decision should be considered as \textit{Joint Accountability} between the healthcare professionals and the AI developing team. 
Adding to this argument, Staszkiewicz et. al. \cite{Staszkiewicz2024legalaccountabilityjoint} argue that making AI developers legally accountable for the AI decisions may lead to institutional conflict and moving against the use of AI. Thus, they advocate for joint accountability of humans and machines.
In our work, we promote for joint accountability between the actors (AI team and the healthcare professionals) as it releases the accountability tensions between the actors, promotes healthy communication, and minimises the chances of scapegoating and blaming.
Moreover, we promote explainability as one important tool that facilitates communication and information sharing between these two actors, and we argue that it should be used for individual decision-making as well as for joint accountability decisions.

\section{Limitations and Future Work}
The main idea of this paper is to declutter the different interpretations of accountability from the regulatory authorities with a top-down approach, with that of the actors with a bottom-up approach. For this purpose, this paper works on a higher-level understanding of the communities and builds on the existing literature. However, this paper does not provide an empirical grounding of the proposed framework, and this is part of the future work.

This paper also opens up the prospect of having joint accountability for the complex ecosystem of AI in the Healthcare environment. With the push and pull effect of bureaucratic accountability and the possibilities of scapegoating and blame-gaming, joint accountability for shared dependencies provides an opportunity for collaboration. However, the criticality of the healthcare scenario, autonomy of the actors and AI, interdisciplinary communication, and many such factors play an important role in how much accountability could be shared among the stakeholders. Moreover, formalising and operationalising this perspective needs deliberate discussions and actions to be taken.

On similar lines, we propose to leverage explainability as a mechanism for facilitating the collaborations. Explainability methods may differ widely based on what is to be explained (\textit{the explanandum}), how it is explained (\textit{the explanans}), when it is explained (post-hoc or transparent), model dependency (agnostic or not), and the types of data \cite{2020_XAI_review_Alejandro,we_need_to_talk_Freiesleben_2023}. Therefore, leveraging explainability depends on a case-by-case basis, based on the need for information exchange and the stakeholders involved. Thus, utilising these methods for the aforementioned purpose is open for research with a lot of flexibility.


\section{Conclusion}
\label{sec:conclusion}
Accountability is often discussed by regulatory bodies in a vertical, top-down approach, as a compliance mechanism, where the authority in control may pose questions, and the actors are required to provide justifications for their actions. This assumes that responsibilities are clear and well-known and a single person can be held accountable, whereas often with AI, the situation is fuzzier as multiple organisations work independently while sharing dependencies. Hence, in practice, 
these accountability mechanisms take various forms due to the varying nature of the work, enforcement methods and authorities, and the degree of implications on various actors. In the literature, it often remains implicit how and why a particular method may contribute to accountability, and their arguments highlight the multifaceted nature of accountability. 
This gives rise to ambiguity around the term and inconsistencies in implementing accountability mechanisms in practice. 

To address this issue, we propose an accountability framework for healthcare AI systems derived from the fundamental definitions originating from the administrative and governance domains. The framework illuminates the actors in the healthcare AI system, their interdependent processes, and bridges the gap between the "what" and the "how" by positioning different mechanisms in practice with the compliance measures. Often, from a practitioner's perspective, accountability measures take various forms, such as managing responsibilities, complying with the regulations, performing audits, quality checks for products, and following certain standards in order to provide justifications for actions and ensure safe working practices. Currently, these practices are governed by individual organisations, even when they have shared dependencies.
As highlighted by our framework, we reflect on three challenges of the healthcare AI system: 1) independent authorities in control and unclear handover processes within the organisations, 2) unclear accountability structure for shared dependencies, and 3) interdisciplinary miscommunication. In addition, we propose a three-tier accountability structure for the actors based on their conduct. Although in practice, most emphasis is given on the product and process level accountability, we also highlight the importance of decision-level accountability. We emphasise that even though the final decisions in a healthcare AI setting might be made by the healthcare professional, the accountability for shared dependencies should be dealt with in a joint manner. This promotes collaboration instead of the blame-gaming found in a bureaucratic accountability structure. 

\section*{Acknowledgments}
This research is partially supported by SPATIAL and ENSURE-6G projects that have received funding from the European Union's Horizon 2020 research and innovation programme and MSCA, under grant agreement No.101021808 and No. 101182933, respectively.
\appendix

\bibliography{aaai25}

\end{document}